# Current LLMs still cannot "talk much" about grammar modules: Evidence from syntax

**Mohammed Q. Shormani, Ibb University, Yemen**
shormani@ibbuniv.edu.ye/https://orcid.org/0000-0002-0138-4793

**Yehia A. AlSohbani, Arab Open University, Saudia Arabia**

## Abstract

We aim to examine the extent to which Large Language Models (LLMs) can "talk" about grammar modules, providing evidence from syntax core properties translated by ChatGPT into Arabic. We collected 44 terms from generative syntax previous works, including books and journal articles, as well as from our experience in the field. These terms were translated by humans, and then by ChatGPT-5. We then analyzed and compared both translations. We used an analytical and comparative approach in our analysis. Findings unveil that LLMs still cannot "talk much" about the core syntax properties embedded in the terms under study involving several syntactic and semantic challenges: only 25% of ChatGPT translations were accurate, while 38.6% were inaccurate, and 36.4.% were partially correct, which we consider appropriate. Based on these findings, a set of actionable strategies were proposed, the most notable of which is a close collaboration between AI specialists and linguists to better LLMs' working mechanism for accurate or at least appropriate translation.

## 1. Introduction

Large Language Models (LLMs) are deep learning Artificial Intelligence (AI) systems including ChatGPT Gemini, Grok, DeepSeek. They are considered a revolution in AI industry. AI is one of the most significant technological advancements of our time. It has attracted tremendous global attention due to its ability to simulate human intelligence through neural systems that can learn from data, logical reasoning, language understanding, and interacting with the environment. Modern AI relies primarily on techniques such as machine learning and deep learning algorithms, which enable it to analyze massive amounts of data, detect specific patterns, and make highly accurate decisions. Today, AI has become a central axis of global competition by improving productivity, automating processes, and creating new jobs in fields such as data engineering and systems analysis (see e.g. McShane & Nirenburg, 2021). AI is also a key driver of innovation and will continue to transform how humans live and work in the near future. However, its widespread adoption raises many challenges, including the ethics of AI use, privacy concerns, and the possibility of replacing humans in professional roles.

The term *Artificial Intelligence* refers to the creation, modeling, and/or production of computer-based intelligence similar to "human intelligence." It can be defined as the science and engineering of enabling computers to exhibit intelligence, where intelligence refers to the ability to perceive, reason, learn, and act in a given environment to achieve goals (Shormani, 2024c). Since its beginnings in the early 1950s, AI has developed remarkably (Minsky, 1961; Turing, 1950; Linzen & Baroni, 2021; Gulordava et al., 2018), leading to the design of computational models that can carry out complex tasks. This progress stems from the integration of Neural Network Algorithms (NNAs), which function intelligently like the human brain. They operate through mechanisms resembling human cognition, known as deep LLMs, which perform tasks with high precision (Gulordava et al., 2018; Linzen & Baroni, 2021; Peng et al., 2023). These models have been developed by AI experts and computer programmers using programming languages that allow them to learn any linguistic phenomenon, be it phonological, morphological, syntactic, or semantic, provided they are trained on sufficient and effective data, in much the same way as a child or adult learns a second language (Shormani ,2024b, Shormani & AlSohbani, 2025).

A child, for instance, acquires its first or second language through its innate Faculty of Language (FL). This faculty consists of a Computational System of Human Language ($C_{HL}$) and the lexicon. The $C_{HL}$ is made up of a set of computational rules and algorithms that can "generate" an infinite number of sentences from a finite set of words and rules. The first language also contains Universal Grammar), which comprises principles and parameters. Principles are common across all natural languages, while Parameters are specific to individual languages (see e.g. Chomsky, 1981; Cook, 1983, 1988; Long, 2003; White, 2000, 2003; Clark, 2009; Shormani, 2014, 2025a & b). This development in AI technologies, tools, and models has significantly advanced the field of Machine Translation (MT). It has evolved from Rule-based Machine Translation (RMT), which focused on syntactic structures, to Statistical Machine Translation (SMT), and finally to Neural Machine Translation (NMT). The latter introduced substantial improvements, most notably the integration of NNAs in AI, which paved the way for the emergence of many deep learning models such as ChatGPT, DeepSeek, and Grok-3. With these advances, the profession of translation has become much easier and more accessible across various domains, as massive amounts of texts and documents can now be translated at remarkable speed and with improved quality, specifically when using neural models. The same applies to related tasks such as subtitling, dubbing, and beyond.

In this study, we address an important aspect of LLMs' translation, selecting ChatGPT as one of these models to test its ability to translate 44 English core syntax terms into Arabic. These terms were collected from books and scholarly research articles published in reputable journals. We then translated them using ChatGPT and compared the output with human translation in terms of convergence and divergence. Through this comparison, the study seeks to assess the extent to which LLMs, specifically ChatGPT can translate syntax core terminology and capture syntactic properties with their nuances. It also aims to identify the challenges ChatGPT faces in translating such terms and to propose suitable solutions that may help resolve or minimize these issues. This article is thus set up as follows. Section 2 outlines the theoretical framework, highlighting key notions from generative syntax and AI translation studies. Section 3 describes the methodology employed, including data collection, translation procedures, and the analytical approach. Section 4 presents and discusses the results of the study. Section 5 concludes the article, stating the study limitations, and suggestions for future research.

## 2. Theoretical Framework

### 2.1. Generative syntax: An historical overview

Generative syntax has undergone numerous developments since its emergence in Chomsky's seminal work *Syntactic Structures* (1957). In this work, Chomsky laid the foundations of generative syntax. Since then, the theory has evolved rapidly, and its main developmental stages can be summarized as follows. According to Shormani (2024a), the first stage is the *Standard Theory*, in which Chomsky proposed the concept of the *kernel sentence*. In this principle, a kernel sentence is defined as irreducible sentence, generated from simple structures by *Phrase Structure Rules*. The second stage is the *Extended Standard Theory*, where semantic roles were integrated into *X-bar Theory*. Here, Chomsky argued that *Transformation Rules* could be applied without explicitly specifying semantic roles, specifically in deep structures such as reference and coreference, and that abstract elements could be introduced. These developments paved the way for what later known as the *Government & Binding Theory* (Chomsky, 1981, see also Shormani, 2024a, 2025a).

One major outcome of these developments was the emergence of a new framework or program referred as *Principles and Parameters* (P&P). The P&P framework marked a genuine revolution in theorizing generative syntax phenomena, as they entered the realm of biology, being treated as a scientific discipline on par with mathematics, physics, and biology. This may be referred to as the third stage. Chomsky's central concern in P&P was to seek scientific and rational answers to fundamental questions of linguistic inquiry (see also Chomsky, 1991, et seq). Thus, one of the most significant changes in P&P was the incorporation of biology into the study of syntax phenomena. It introduced the idea that humans possess an FL, a mental organ responsible for acquiring, processing, and interpreting language. The FL contains Universal Grammar, composed of principles and parameters. Principles are common to all languages, whereas parameters are specific to individual languages. For example, *Subject Principle* states that "every sentence must have a subject." This applies in all languages, but the subject may be either overt or covert (implicit/*pro*). The overt case occurs in languages such as English, Hindi, and French, while the covert case appears in languages such as Arabic, Greek, and Irish. This is called *Null- subject Parameter* (see e.g. Shormani, 2017, 2024a).

However, the P&P framework was highly complex, overloaded with theories, assumptions, and hypotheses. These complications led to the emergence of the *Minimalist Program* (MP), the fourth stage in the evolution of generative syntax. Put differently, the complexities of P&P urged Chomsky to propose the MP (1993-2021), which rests on the notion that human language is not as complex as earlier generative theories assumed. The MP, specifically its most recent version known as the *Phase Theory*, posits that language consists of two levels: ***sound*** and ***meaning*** (Chomsky, 2001, et seq; Citko, 2014). Between these two levels, the human mind computes, processes, and interprets any linguistic expression s/he speaks or spoken to him/her. In MP, language is regarded as "an optimal solution to the condition of legibility" (Chomsky, 2000, p. 97).

Throughout these four stages, numerous technical terms emerged, each carrying significant theoretical and conceptual weight. These English terms have posed real challenges to Arab students of generative syntax, particularly those who find themselves compelled to translate these terms into Arabic to grasp their nature. Consequently, many of them resort to LLMs such as ChatGPT. In this study, we aim to examine ChatGPT's potential to translate such terms, bearing in mind that it was trained on massive (internet) data (up to 2021) and can now browse the web (particularly ChatGPT-5) to retrieve and reformat information as needed. Recent years have also witnessed the emergence of other LLMs comparable to ChatGPT, such as DeepSeek and Grok-3 (see also Shormani & Al-Samki, 2025).

### 2.2. LLMs and translation

Translation is one of the oldest means humans have utilized to transmit knowledge and foster intercultural communication, being not merely about transferring the meaning of words but also involves creativity and innovation (see e.g. Newmark, 1988; Bassnett, 2014; Shormani, 2020). In the Arab context, translation dates back to ancient civilizations such as the Sumerian and Egyptian. In modern times, translation has moved beyond the purely linguistic level to encompass cultural, literary, and semiotic dimensions (see e.g. Lee, 2023). This has given rise to new concepts such as *translation editing* in literary works, which highlights the translator's role as a creative re-author who reshapes texts in accordance with the target culture (see e.g. Shormani, 2020). Literary translation, specifically poetry translation, remains one of the most challenging types (see also Jakobson, 1959). Translation today also engages deeply with cultural challenges, as seen in the translation of proverbs. For instance, English proverbs (Shormani, 2020) require semiotic interpretation that goes beyond the "common" or surface meaning in order to convey the cultural and symbolic depth of the source text.

MT was the first attempt to use some form of AI in translation, attracting specialists in computational linguistics (CL), Natural Language Processing (NLP), and even general linguistics. It involves producing and reading linguistic texts via computers, as well as employing methods, extracting linguistically outstanding and reliable information from these texts (Brown et al., 1990). MT has been advanced considerably, particularly in recent years, evolving from rule-based systems to statistical ones and finally to neural networks (de Coster et al., 2023; Dugonik et al., 2023). Today, MT is applied in numerous fields, including film production, subtitling, lexicography, and dictionary-making (Krings, 2001; Baniata, et al., 2021; Zhang & Torres-Hostench, 2022). AI constitutes the foundation of rapid developments in machine translation, having made significant contributions to the translation industry. It has revolutionized this field by modeling complex algorithms capable of learning and translating texts across languages. Early MT systems were based on rule-based translation (RMT), which operated primarily on syntax structures. Later, statistical machine translation emerged, followed by NMT (Koehn et al., 2003; Forcada et al., 2011). The central idea behind RMT is that a word in Language 1 has an equivalent in Language 2, and the machine simply substitutes one for the other according to syntax rules. The shortcomings of RMT motivated the emergence of SMT in the early 1990s, during the rise of "artificial intelligence" in the computer industry (Pollack, 1990). SMT is based on the notion that words may have multiple meanings, and the machine identifies the most probable equivalent by relying on bilingual corpora. A bilingual corpus functions as a bilingual lexicon, with WordNet being a prime example. This mechanism underlies Google Translate, where the system selects the best match or meaning for a word in L1 from L2, depending on its recurrence in bilingual corpora. SMT represented a considerable advancement in MT and remained dominant for a long period, during which various SMT models were developed and continuously refined. However, SMT outputs were often unsatisfactory, as seen in Google Translate.

With major advances in technology and AI, MT entered a new era through the integration of neural networks and deep learning architectures. NMT systems rely primarily on neural network algorithms, enabled by the rapid progress in AI technologies (Peng et al., 2023). The central principle here is to make computers "think" or perform tasks similar to humans. The incorporation of NNAs into AI began around 2010 with the rise of deep Learning LLMs, such as ChatGPT. Since then, numerous LLMs have been developed, including hybrid SMT–NMT models (Dugonik et al., 2023), multi-NMT models (Huang et al., 2023), transformer-based architectures (Baniata et al., 2021), human-in-the-loop post-editing models (Formiga et al., 2015), and human–AI collaborative frameworks (Li et al., 2023). Supported by NLP and AI, NMT has significantly advanced the translation industry, attracting researchers across fields such as linguistics, computer science, engineering, programming, CL, and NLP. These developments have yielded cutting-edge applications like ChatGPT, DeepSeek, and Grok-3.

ChatGPT, developed and launched by OpenAI, is among the prominent LLMs designed for multiple purposes, including translation. It is also capable of generating content across domains such as text, images. Beyond these functions, ChatGPT has been integrated into scientific research, ranging from data processing and hypothesis generation to public communication (Ray, 2023, p.121). As a large neural language model, ChatGPT represents a deep learning system designed to produce translations of higher accuracy than other MT tools, such as Google Translate (Lee, 2023; van Dis et al., 2023).

Thus, the real era of implementing a sophisticated output of AI in translation is perhaps with incorporating NNAs in AI systems, coming up with large language models. LLMs become integral to the daily work of many practitioners including translators, academics, students, physicians, and many others. In translation sphere, LLMs including ChatGPT have rendered translation into a computational process, enhancing both its speed and its productivity (Dugonik et al., 2023; Lee, 2023). AI has also played a major role in the success of translation as industry, particularly through the application of its latest tools and models, which have redefined its scope and made it more efficient than ever. Today, AI translation is embedded in numerous fields, with AI serving as its backbone. The profession of translation has thus undergone a profound shift, transforming from a purely human craft into a "computational" practice. This transformation or, say, reconfiguration stems mainly from the rapid advancements in technology, the internet, and LLMs. However, these developments have also given rise to challenges, particularly those relating to linguistic and cultural differences across languages, in addition biases, ethics (Shormani, 2024c, 2025a).

Several studies have investigated ChatGPT's potential to translate various types of texts. For example, Shormani (2024c) examined ChatGPT's capacity in religious translation, focusing on the precise rendering of Arabic oath expressions into English while retaining their religious and cultural nuances. The study analyzed 40 commonly used oath expressions in Arabic, comparing ChatGPT's output with human translations. Findings reveal significant gaps in the AI's performance, particularly its failure to capture cultural-religious aspects accurately, its use of inappropriate equivalents for oath markers, and tendencies to omit, add, or distort meanings. The study concludes that despite its advanced linguistic capabilities,

ChatGPT requires substantial improvements in handling Arabic oath texts and recommended training the model on more culturally and religiously diverse datasets.

Alwagieh and Shormani (2024) examine the effectiveness of ChatGPT in translating Arabic free-verse poetry into English. They analyze selected poems from Wahib's *Hulm 'ala Nasiat Aljabin* (*A Dream on the Corners of Nostalgia*) and compare translations produced by ChatGPT with human translations. Using a qualitative comparative approach, the study evaluates differences in translation strategies and the extent to which the AI output deviates from human translation. The findings show that ChatGPT can produce grammatically natural and generally understandable English translations, indicating some success in conveying the basic meaning of the poems. However, the system tends to rely heavily on literal translation, word-for-word rendering, and occasional paraphrasing additions, whereas human translators employ more sophisticated strategies such as adaptation, recreation, paraphrasing, and transliteration to convey poetic meaning and cultural nuance. Consequently, ChatGPT's translations often neglect aesthetic and stylistic elements, including imagery and poetic effect, leading to various translation inaccuracies. The study concludes that while AI tools may assist in preliminary translation, human translators remain essential for preserving the artistic and cultural qualities of Arabic poetry.

Another study conducted by Larroyed (2023) explores how recent LLMs such as ChatGPT may affect patent translation practices and language policies in Europe. The study adopts an interdisciplinary approach combining legal analysis, literature review, and a preliminary empirical evaluation of ChatGPT's translation performance. Using the LISA QA translation quality assessment framework, the author evaluates translations of a sample of patent texts and compares ChatGPT's output with the existing patent translation system "Patent Translate." Findings suggest that ChatGPT can produce highly fluent and syntactically well-formed translations and, in some cases, achieves translation quality comparable to existing machine-translation systems. However, specialized patent terminology may still present challenges, and the legal reliability of machine-generated translations remains uncertain. The article therefore argues that advances in AI translation technology require reconsideration of current European patent language regimes, particularly those established under the European Patent Convention. While AI tools can improve accessibility to patent knowledge, the study emphasizes that human oversight and regulatory adaptation are necessary to ensure accuracy and legal validity in patent translation.

Additionally, Shormani & AlSohbani (2025) conducted a large-scale scientometric and thematic analysis of 9,836 research articles from WoS, Scopus, and Lens databases. The scientometric analysis examined research groups, subject categories, keywords, and institutional affiliations, while the thematic analysis reviewed 18 purposively sampled articles to evaluate objectives, methodologies, findings, and future recommendations in AI and translation research. Results showed that AI's role in translation was limited in its early stages, when it relied on rule-based and statistical MT approaches, which yielded unsatisfactory results. However, the emergence of neural algorithms and deep learning models like ChatGPT has significantly improved translation quality. However, persistent challenges remain in translating languages with low-resource data, multi-dialectal varieties, and complex sentence structures, as well as in addressing cultural and religious contexts. The study recommended further research into developing more accurate and adaptable AI-based translation systems.

A final study that could be highlighted here was conducted by Lee (2023). Lee examines the implications of recent LLMs, specifically ChatGPT for translation theory and professional translation practice. Rather than conducting a purely technical evaluation of translation quality, the study adopts a conceptual and theoretical perspective grounded in posthumanist thought to explore how AI challenges traditional views of translation and the role of human translators. The article argues that LLMs have reached a level of sophistication where they can sometimes produce translations comparable to human translation. This development significantly reshapes translation workflows and raises questions about the identity and future role of translators in an AI-mediated environment. Drawing on posthumanist and distributed cognition frameworks, Lee proposes that translation should no longer be viewed as a purely human activity but as a distributed process involving both human and technological agents. In this perspective, AI systems function as a kind of *cognitive extension* or "prosthesis" that augments the translator's capabilities rather than replacing them. The article therefore calls for a redefinition of translation competence that incorporates technological collaboration and broader multimodal communication skills.

Based on this brief overview of prior research, the present study offers a significant contribution to understanding the potential of AI-driven translation, particularly through ChatGPT, while also paving the way for future investigations. Specifically, it seeks to answer the following questions:

1. To what extent can LLMs, specifically ChatGPT capture syntax nuances when translating English syntax terms into Arabic?
2. What challenges does ChatGPT encounter in rendering these syntactic details?
3. What are the solutions that can be proposed to overcome such challenges?

## 3. Methodology

### 3.1. Study data

We selected 44 core and widely used syntax terms from previous studies in generative syntax and from our own experience as a specialist in the field. The terms reflect the historical development of generative syntax, from early stages to the MP, thereby ensuring comprehensiveness. They also represent concepts taught at the (under)graduate levels, such as *pro*, *Case*, and *Government & Binding Theory*. The dataset also includes compound terms containing hyphens, such as *X-Bar Theory*, in order to examine ChatGPT's handling of structurally diverse terminology. The terms were translated using ChatGPT-5 (the free version) and compared with academically reliable human translations. The study was conducted on September 28, 2025. ChatGPT-5 was chosen over earlier models (GPT-2, -3, -3.5, -4o) because it is the latest release and integrates several improvements over its predecessors. Fig 1 below presents a sample of ChatGPT translations.

ChatGPT

Free offer ×

1. **pro** – برو (ضمير مستتر غير ملفوظ يُفترض وجوده)
2. **Case** – الحالة الإعرابية
3. **Goal** – الهدف (أو الغاية في علاقة الجملة)
4. **Probe** – المجسّ (أو الكاشف النحوي)
5. **Passive** – المبني للمجهول
6. **Lexicon** – المعجم
7. **Specifier** – المحدِّد
8. **Principles** – المبادئ
9. **Projection** – الإسقاط
10. **Nominative** – الرفع (أو الحالة الرفعية)

***Fig 1: A Sample of ChatGPT translation***

The study employed a descriptive, analytical, and comparative approach, analyzing and discussing the results using both qualitative and quantitative methods. In addition, it relied on the micro-level approach (see for example Fairclough, 2003), which focuses on analyzing the linguistic aspects of discourse. In this study, we applied these scientific methods to analyze ChatGPT translations, describe them, and then compare them with human translations of generative syntax terms.

## 4. Results and discussion

In this section, we first present and discuss the results obtained. Table 1 displays a sample of the results.

**Table 1: GS term, ChatGPT translation, and human translation**

| No. | GL term | ChatGPT translation | Human translation |
|---|---|---|---|
| 1 | *pro* | برو (ضمير مستتر غير ملفوظ يفترض وجوده) | الضمير المستتر |
| 2 | Case | الحالة الإعرابية | الحالة الإعرابية |
| 3 | Goal | الهدف (او الغاية العلاقة في الجملة) | الهديفة |
| 4 | Probe | المجسّ (أو الكاشف النحوي) | السبيرة |
| 5 | Passive | المبني للمجهول | المبني للمجهول |
| 6 | Lexicon | المعجم | المعجم |
| 7 | Specifier | المحدِّد | المحدد |
| 8 | Principles | المبادئ | المبادئ |
| 9 | Projection | الإسقاط | انشطار |
| 10 | Nominative | الرفع (أو الحالة الرفعية) | حالة الرفع |
| 11 | Parameters | المعلمات | الوسائط |
| 12 | Accusative | النصب (أو الحالة النصبية) | حالة النصب |
| 13 | Argument Structure | البنية الحجاجية (أو بنية الحجج) | بنية الموضوع |
| 14 | Generative Grammar | النحو التوليدي | النحو التوليدي |
| 15 | X-Bar Theory | نظرية X-bar | نظرية س- |
| 16 | Syntactic Adequacy | الكفاية النحوية | الكفاية النحوية |
| 17 | Phrase markers | محددات العبارة أو علامات البنية التركيبية | أشجار نحوية |
| 18 | Deep Structure | البنية العميقة | البنية العميقة |
| 19 | Syntactic Interpretation | التفسير النحوي | التفسير النحوي |
| 20 | Syntax-discourse interface | واجهة النحو والخطاب | وجيهة النحو-خطابية |

Table 2 above shows that ChatGPT translations could be studied in terms of two categories, viz., *Translation strategy* and *Translation accuracy*. These are discussed in the following (sub)sections.

### 4.1. Translation strategy

This type includes the following subcategories: i) Translation by addition, ii) Morphological derivation, and iii) Arabization and Englishization

As for the first category, i.e. *Translation by Addition,* it refers to cases where ChatGPT translates a term by adding a phrase or explanation in parentheses or using “or”. For example, *pro* was translated as برو (ضمير مستتر غير ملفوظ يفترض وجوده) while the human translation الضمير المستتر. Here, ChatGPT shows uncertainty, adding an explanatory phrase in parentheses, which partially approximates the human translation. In generative syntax, *pro* refers to a null subject pronoun in pro-drop languages such as Arabic and Hebrew, unlike English or French where null subjects are almost disallowed. The first case in point here is *Nominative* which was rendered as الرفع (أو الحالة الرفعية)”, where ChatGPT adds (أو الحالة الرفعية) which is some sort of explanatory and close to the intended meaning. However, human translates it as “حالة الرفع”. Likewise, “Accusative” was rendered by ChatGPT as النصب (أو الحالة النصبية), where ChatGPT adds as (أو الحالة النصبية)”, which is inaccurate. A final example that can be stated here is translating the term “Left Periphery” as المحيط الأيسر (أو البنية النحوية). Such cases reveal that ChatGPT often compensates for uncertainty by inserting additional wording. However, this phenomenon demonstrates its ability to explain or paraphrase, which indicates potential for improvement.

As for the second subcategory, viz., *morphological derivation*, and as the name suggests, differences arise from how ChatGPT handles derivation compared to human translation. Human translators often derive adjectives or nouns beyond the given form. For instance, *Sound-meaning Interface*, which was translated by human as وجيهة الصوت-دلالية while ChatGPT translates it as واجهة الصوت والمعنى, and *Articulatory-conceptual System,* which was translated by human as نظام النطق-مفاهيمي, but ChatGPT translates it as النظام النطقي-التصوري. Additionally, the human translator derived مفاهيمي (adjective form of مفهوم), whereas ChatGPT produced التصوري, which is not accurate syntactically. A similar issue occurs with *Linearization* which was rendered by ChatGPT as الخطية while human translates it as التخطيطية. The latter derives from the verb خطط ‘to mark lines,’ matching the intended meaning in generative syntax, where *linearization* refers to the process of arranging syntactic units into a linear order (see e.g. Kayne, 1994).

By *Arabization and Englishization*, we mean a process or a type of translation which takes the form of transliteration. In our context, Arabization appears in cases like *pro* rendered as برو. Englishization appears in terms such as *X-bar Theory* rendered as نظرية *X-bar* by ChatGPT instead of the conventional نظرية س-, which has long been used in Arabic generative syntax (see e.g. Al-Rahali, 2003; Ghulfan et al., 2010). Although the data exists in ChatGPT’s training corpus, the model retained the English form, reflecting an alienating strategy. Or, perhaps the term نظرية- س is not “picked up” by ChatGPT, and this unveils some kind of bias concerning Arabic language.

### 4.2. Translation accuracy

This type includes the following subcategories: 1) Correct translation, 2) Appropriate translation, and 3) Completely incorrect translation. To start with, *Correct translation* refers to cases where ChatGPT's output exactly matches human translation, both lexically and scientifically. For example, translating the term *Passive* as المبني للمجهول, *Lexicon* as المعجم, Generative grammar as النحو التوليدي, and *Transformational Grammar* as النحو التحويلي. Such terms account for about 22.5% of the total (see Table 3). Additionally, *Appropriate translation* reflects ChatGPT translation as close but not fully accurate translations. For instance, *Minimalist Program* was rendered by ChatGPT as البرنامج التبسيطي. While البرنامج is correct, التبسيطي is misleading, as it suggests stylistic simplicity rather than the intended syntactic sense of الأدنوي. Other examples include *Interface Level Legibility* which was rendered as قابلية القراءة في مستوى الواجهة while human translated it as مقروئية مستوى الوجيهة.

The third subcategory in accuracy of translation is *completely incorrect translation*. What is meant by this subcategory is those cases where ChatGPT's output diverged entirely from the intended meaning. For example, the term *Government was rendered as* الحكم, a political term unrelated to syntax. Another example is *Probe* rendered as المجس, which is not correct, as is clear from the human translation, viz., السبيرة. السبيرة is the accurate term here due to the fact that *probe* is the active ingredient during the derivation process, and which "search" the search space for matching features. Translating *Probe-goal Relation* is another example exemplifying this type of ChatGPT translation. This tern was rendered by ChatGPT as العلاقة المجس بالهدف, while human renders it as العلاقة بين السبيرة والهديفة. A finally example that can be concretized here is *Discourse Configurational Language* which was rendered by ChatGPT as اللغة ذات التهيئة الخطابية, but human translator renders it as لغة بناء الخطاب.

We also intended to use statistics to provide a precise assessment of ChatGPT's performance in translating the selected generative syntax terms by calculating the number and percentage of correct, appropriate, and incorrect translations. Table 3 below presents the quantitative analysis of ChatGPT translation of the 44 terms.

**Table 3: ChatGPT's output in statistics**

| Category | Frequency | % |
|---|---|---|
| Correct translation | 11 | 25 |
| Appropriate translation | 16 | 36.4 |
| Incorrect translation | 17 | 38.6 |
| Total | 44 | 100 |

As shown in Table 3, the accurate translations produced by ChatGPT that match human translation in terms of syntax core interpretation amounted to only 11 cases, i.e. 25% of the sample. The near-correct or appropriate translations, which contained some accuracy but deviated from the human rendering are 16 cases, viz., 36%. Finally, the number of incorrect translations, which failed to convey the intended syntactic core properties/nuances are 17 cases, precisely 38.6% of the total terms under study.

To recapitulate, it seems that the ChatGPT incorrect translation often result from literalism or insufficient exposure to specialized generative syntax terminology, leading to misrepresentation of the concepts. The qualitative and quantitative results of ChatGPT translations of generative syntax terms, compared to those of human, revealed significant syntactic, lexical, and semantic differences. These results highlight the challenges LLMs such as ChatGPT still lag behind translating English generative syntax terms into Arabic. It can be said that full reliance on machine translation, even with advanced LLMs like ChatGPT, remains insufficient without human intervention, making a collaborative approach between AI and human expertise essential.

## 5. Conclusions and limitations

This study tested LLMs, specifically ChatGPT's potential to translate English syntax core properties into Arabic. 44 terms were collected from generative syntax literature and translated into Arabic first by human translators and then by ChatGPT. The translations were subsequently analyzed and compared. The results showed that only 25% of ChatGPT translations were accurate, while 38.6% were incorrect, and 36.4% were only partially correct, which we consider appropriate. These findings suggest that while ChatGPT is a useful LLM, it cannot yet be fully relied on for specialized translation in this field, at least currently, though further training on such terminology may improve future performance. Accordingly, it is not advisable for students of generative syntax to depend solely on LLMs such as ChatGPT for translating these terms without human involvement in post-editing or revision. This study concludes that translating English generative syntax terms into Arabic using ChatGPT still requires substantial human intervention. Such intervention is vital to correct errors, address inadequacies, and prevent distortions of intended meaning, which could otherwise lead to confusion or ambiguity among students of generative syntax. Human translators can thus benefit from ChatGPT's speed while adding the essential "human touch" needed for high-quality translations (see also Krings, 2001; Daems et al., 2017; Balling et al., 2014).

Additionally, to overcome these limitations and enhance its translation performance, ChatGPT, and LLMs in general, could be further trained on the gaps identified in this study. collaboration between AI specialists and linguists is essential in developing actionable strategies, structured, operational resources that guide translation models toward consistent and theoretically accurate terminology. These strategies can include curated bilingual terminological databases of generative syntax, annotated corpora of Arabic linguistic literature, and standardized glossaries aligned with the generative enterprise starting from its inception, targeting works (or earlier) like Chomsky *Syntactic Structures* through P&P and finally Minimalism (see Chomsky, 1957-2021). By compiling authoritative Arabic equivalents for core syntactic concepts starting from Sibawieh to the present time scholars. In addition, these strategies may incorporate metadata on morphological patterns, syntactic and semantic structures, and preferred translation strategies (see also Shormani & AlSohbani, 2025). Such strategies can significantly enhance the reliability, terminological consistency, and scholarly adequacy of LLMs' translation in specialized fields such as generative syntax, given that these models are developed every single day.

However, the study has some limitations: i) the sample of terms involved was limited to 44 generative syntax terms, which may not fully capture the diversity and complexity of specialized syntactic terminology in English and Arabic, ii) the analysis relied primarily on a single AI model, ChatGPT, without comparing its performance to other AI translation tools or neural machine translation systems, which could provide a broader perspective on AI capabilities, and iii) the study focused exclusively on the translation of isolated (single) terms without considering context-rich discourse or sentence-level usage, which may influence translation accuracy and appropriateness. Future research could expand the dataset to include a wider range of syntactic terms, incorporate multiple AI translation models such as DeepSeek or Grok-3 for comparative evaluation, examining the performance of AI in translating full syntactic structures or contextually embedded examples to better assess its utility in longer contexts, perhaps within a discourse.

**Declarations**

**Funding**
This research received no internal or external funding.

**Clinical Trial Number**
Not applicable

**Clinical trial number**
Not applicable.

**Consent to Publish**
Not applicable.

**Competing Interests**
The authors have no competing interests to declare.

**Consent to Participate**
Not applicable.

**Ethics Declaration**
Not applicable.